\documentclass[runningheads]{llncs}
\usepackage{amsmath}
\usepackage{graphicx}
\usepackage{lipsum}
\usepackage{mwe}
\usepackage[dvipsnames]{xcolor}
\usepackage{multirow}
\usepackage{cite}
\usepackage{amsmath}
\usepackage{tabu}
\usepackage{subfig}
\usepackage[misc]{ifsym}
\newcommand{\etal}{\emph{et~al.}}

\newcommand{\corrauth}{\textsuperscript{(\Letter)}}

\usepackage{color}
\newcommand{\thickhline}{%
    \noalign {\ifnum 0=`}\fi \hrule height 1pt
    \futurelet \reserved@a \@xhline
}
\graphicspath{{./figures/}}

\begin{document}
\title{Sensorless Freehand 3D Ultrasound Reconstruction via Deep Contextual Learning}
\titlerunning{Sensorless Freehand Ultrasound Reconstruction}
\author{Hengtao Guo\inst{1} \and
Sheng Xu\inst{2} \and
Bradford Wood\inst{2} \and
Pingkun Yan\inst{1}\corrauth}
\institute{Department of Biomedical Engineering and Center for Biotechnology and Interdisciplinary Studies, Rensselaer Polytechnic Institute, Troy, NY 12180, USA\\
\email{yanp2@rpi.edu}\\
\and
National Institutes of Health, Center for Interventional Oncology, Radiology \& Imaging Sciences, Bethesda, MD 20892, USA}
%}
%
\maketitle              % typeset the header of the contribution
\begin{abstract}
% \redb{Rewrite after you finish the paper.}
Transrectal ultrasound (US) is the most commonly used imaging modality to guide prostate biopsy and its 3D volume provides even richer context information. Current methods for 3D volume reconstruction from freehand US scans require external tracking devices to provide spatial position for every frame. In this paper, we propose a deep contextual learning network (DCL-Net), which can efficiently exploit the image feature relationship between US frames and reconstruct 3D US volumes without any tracking device. The proposed DCL-Net utilizes 3D convolutions over a US video segment for feature extraction. An embedded self-attention module makes the network focus on the speckle-rich areas for better spatial movement prediction. We also propose a novel case-wise correlation loss to stabilize the training process for improved accuracy. Highly promising results have been obtained by using the developed method. The experiments with ablation studies demonstrate superior performance of the proposed method by comparing against other state-of-the-art methods. Source code of this work is publicly available at \url{https://github.com/DIAL-RPI/FreehandUSRecon}.
%on the volume reconstruction task show that: (1) 3D convolutional layers can reveal the spatial information between nearby frames; (2) The attention block helps the network focusing on areas with correlation; (3) The case-wise correlation loss can make the network more sensitive to the rapid change in probe's moving trajectory. 


\end{abstract}

\section{Introduction}
% (Problem definition) 

%\begin{figure}[t]
%\centering
%\includegraphics[width=\textwidth]{figures/flow.pdf}
%\caption{\cyan{3D ultrasound volume reconstruction from freehand 2D scans through a chain of transformations.} } 
%\label{illustration}
%\end{figure}

Ultrasound (US) imaging has been widely used in interventional applications to monitor and trace target tissue. US possesses many advantages, such as low cost, portable setup and the capability of navigating through patient in real-time for anatomical and functional information. Transrectal ultrasound imaging (TRUS) has been commonly used for guiding prostate cancer diagnosis and can significantly reduce the false negative rate when fused with magnetic resonance imaging (MRI)~\cite{siddiqui2015comparison}. However, 2D US images are difficult to be registered with 3D MRI volume, due to the differences in not only image dimension but also image appearance. In practice, a reconstructed 3D US image volume is usually required to assist such interventional tasks.

A reconstructed 3D US imaging volume visualizes a 3D region of interest (ROI) by using a set of 2D ultrasound frames~\cite{mohamed2019survey}, which can be captured by a variety of scanning techniques such as %2D array transducer~\cite{fenster2001three} \redb{2D array gets 3D volume directly}, 
mechanical scan~\cite{daoud2015freehand} and freehand tracked scan~\cite{wen2013accurate}. Among these categories, the tracked freehand scanning is the most favorable method in a number of clinical scenarios. For instance, during a prostate biopsy, the freehand scanning allows clinicians to freely move the US probe around the ROI and produces US images with much more flexibility. The tracking device, either an optical or electro-magnetic (EM) tracking system, helps to build a spatial transformation chain between the imaging planes in the world coordinate for 3D reconstruction.
%, as shown in Fig.~\ref{illustration}. 

% (Classical sensorless concepts) 
US volume reconstruction from sensorless freehand scans takes a step further by removing the tracking devices attached to the US probe. The prior research on this was mainly supported by the speckle decorrelation~\cite{chen1997determination,tuthill1998automated}, which maps the relative difference of position and orientation between neighboring US images to the correlation of their speckle patterns, \emph{i.e.} higher the speckle correlation, lower the elevational distance between neighboring frames~\cite{chang20033}. By removing the tracking devices, such sensorless reconstruction allows the clinicians to move the probe with less constraint without the concerns of blocking tracking signals. In addition, it also reduces the hardware cost.
%Along this path, Laporte \etal~\cite{laporte2011learning} proposed to estimate the elevational distance by learning its relation to inplane image statistics. Afsham \etal~\cite{afsham2015nonlocal} proposed to use a dedicated non-local means filtering to denoise the US frames for the informative speckle patterns extraction. These successes have shown that speckle correlated regions are crucial for estimating the relative transformation between neighboring frames. Paying more attention to these speckle-rich regions can boost the reconstruction performance.
Although the speckle correlation carries information of the relative transformation between neighboring frames, relying on the decorrelation alone renders unreliable performance \cite{laporte2011learning,afsham2015nonlocal}.

In the past decade, deep learning (DL) methods based on convolutional neural networks (CNN) have been identified as an important tool for automatic feature extraction. In the field of US volume reconstruction, a pioneer work carried out by Prevost \etal~\cite{prevost20183d} explored the feasibility of using CNN to directly estimate the inter-frame motion between two 2D US scans. A 2D convolutional network takes two consecutive ultrasound frames and a generated optical flow field between them as a stacked input to estimate the relative rotations and translations between these two frames. However, a typical US scanning video contains rich contextual information beyond two neighboring frames. A sequence of 2D US frames can provide a more general representation of the motion trajectory of US probe. Using only two neighboring frames may lose temporal information and thus result in less accurate reconstruction. In addition, optical flow field, which is good at describing in-plane motion, may not help out-of-plane motion analysis. Besides, the prior works on decorrelation suggests that paying more attention to the speckle-rich regions can boost the reconstruction performance, which hasn't been explored.

% (Our motivation, proposal and contribution) 
In this paper, based on the above observations, we propose a novel Deep Contextual Learning Network (DCL-Net) for sensorless freehand 3D ultrasound reconstruction. The underlying framework takes multiple consecutive US frames as input, instead of only two neighboring frames, to estimate the trajectory of US probe by efficiently exploiting the rich contextual information. Furthermore, to make the network focus on the speckle-rich image areas to utilize the decorrelation information between frames, the attention mechanism is embedded into the network architecture. Last but not the least, we introduce a new case-wise correlation loss to enhance the discriminative feature learning to prevent overfitting the scanning style.
%The main contributions of this work are summarized as follows: 
%\begin{enumerate}
%    \item We use deep 3D CNNs taking US video segments as inputs to extract video temporal information, leading to a boosted trajectory estimation accuracy. 
%    \item We apply the attention mechanism by embedding an attention module into the network which helps the model focusing on speckle-rich image areas.
%    \item Additionally, we include a case-wise correlation loss to enhance the discriminative feature learning thus prevents the overfitting of the scanning style. 
%\end{enumerate}

%\blue{The rest of this paper is organized as follows. Section ~\ref{data} shows the characteristic of the studied dataset; Section ~\ref{method} explains our method with regard to three proposed contributions; Section ~\ref{experiments} displays the experimenting results and we conclude this work in Section ~\ref{Conclusion}.}

\section{Data Materials}
\label{data}
% \redb{Describe your dataset first, size, acquisition with EM etc..} 
All TRUS scanning videos studied in this work are collected by an EM-tracking device from real clinical cases. The dataset contains 640 TRUS videos all from different subjects acquired by a Philips iU22 scanner in varied lengths.
% of the  (around 16 - 250 frames) . 
% \redb{16 frames in total? That doesn't look right. Remove those cases.}
%For every subject in our dataset, there is a US scanning video. 
%While the videos have varying number of frames, 
Every frame corresponds to an EM tracked vector that contains the position and orientation information of that frame. We convert this vector to a 3D homogeneous tranformation matrix $M = [R \; T; 0 \; 1]$,
%\begin{equation} \label{trans}
%M=\begin{bmatrix}
%R & T\\ 
%0 & 1
%\end{bmatrix},
%\end{equation}
where $R$ is a 3$\times$3 rotation matrix and $T$ is a 3D translation vector.
% This transformation matrix is captured by an EM-tracking system during a TRUS scanning for the purpose of guiding prostate biopsy. 

The primary task of 3D ultrasound reconstruction is to obtain the relative spatial position of two or more consecutive US frames. 
%By orderly placing all US frames of one video in the imaging coordinate system and applying interpolation, a 3D US volume can be rendered. 
Without loss of generality, here we use two neighboring frames as an example for illustration. Let $I_i$ and  $I_{i+1}$ denote two consecutive US frames with corresponding transformation matrices $M_i$ and $M_{i+1}$, respectively. The relative transformation matrix $M'_i$ can be computed as $M'_i=M_{i+1}M_i^{-1}$.
%\begin{equation} \label{relative}
%M'_i=M_{i+1}M_i^{-1}
%\end{equation}
%$M'_i$ is a typical 4x4 transformation matrix and is irrelevant to the coordinate system, as we are calculating the relative position based on  $I_n$. 
By decomposing $M'_i$ into 6 degrees of freedom $\theta_i={\left \{t_x,t_y,t_z,\alpha_x,\alpha_y,\alpha_z \right \}}_i$, which contains the translations in millimeters and rotations in degrees, we can use this $\theta_i$ computed from EM tracking as the ground-truth for network training. 
%This label denotes the inter-frame motion between two consecutive frames, aka $I_i$ and  $I_{i+1}$.

 \section{Ultrasound Volume Reconstruction} 
\label{method}

%In this section, we introduce our proposed DCL-Net for 3D US reconstruction with sensorless freehand scans. We start with introducing the network's basic design, then show the implementation of the attention module, followed by the calculation of case-wise correlation loss. 

% The attention block is a branch that learns the attention map for localizing informative image areas. In addition to the commonly used mean squared error (MSE) loss, a new loss function containing a correlation loss term to force the network to extract discriminative features is introduced as shown in the lower panel of Fig.~\ref{framework}.} \redb{Tell people how do you organize this section.}

%\subsection{Network Design}

% \redb{Describe your overall network, including those blocks etc.}

\begin{figure}[t]
	\centering
	\includegraphics[width=\textwidth]{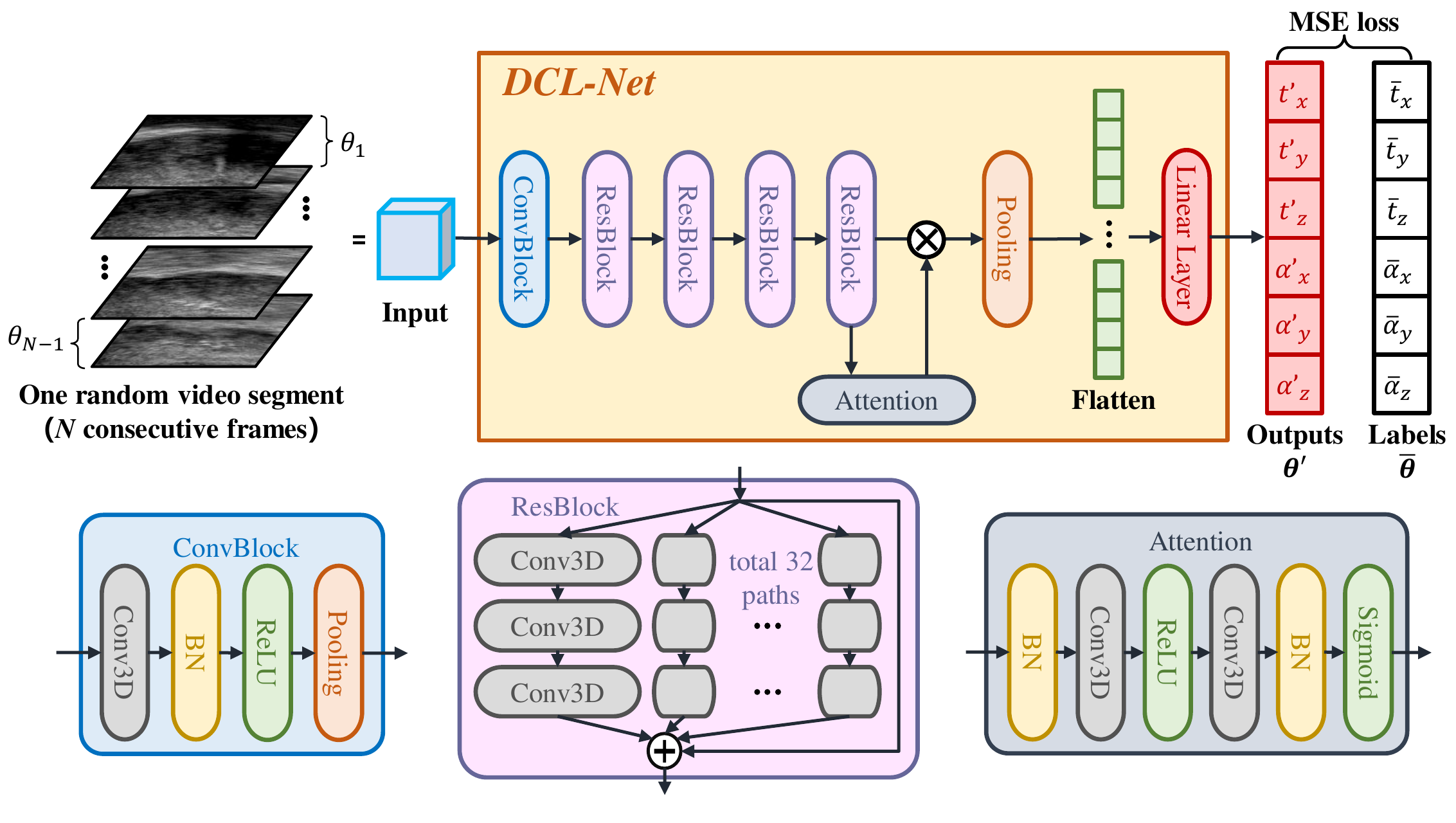}
	\caption{An overview of the proposed DCL-Net, which takes one video segment as input volume and gives the mean motion vector as the output.} 
	\label{fig:framework}
\end{figure}

Fig.~\ref{fig:framework} shows the proposed DCL-Net architecture, which is designed on top of the 3D ResNext model \cite{xie2017aggregated}. Our model consists of 3D residual blocks and other types of basic CNN layers~\cite{he2016deep}. The skip connections help preserve the gradients to train very deep networks. The use of the multiple pathways (cardinalities) enables the extraction of important features. 
% 3D convolution operations are the core contributors to the huge leap of reconstruction performance. 
In our design, 3D instead of 2D convolutional kernels are used, mainly because 3D convolutions can better extract the feature mappings along the axis of channel, which is the temporal direction in our case. Such properties enable the network to focus on the slight displacement of image features between consecutive frames. The network can thus be trained to connect these speckle correlated features to estimate the relative position and orientation.

% \cyan{Given a video segment with $N$ frames with height and width denoted by $H$ and $W$, respectively, the first 3D convolutional layers is designed to have 64 3D kernels sized $7\times7\times3$. The stride equals to 2 on the image plane with 0 paddings and 1 along the image channels (frames) without padding. Such design pattern can capture a relatively large spatial receptive field over three neighboring frames for temporal information extraction.} \redb{How about the other layers? Maybe just remove the entire paragraph.}
%The maximum of a 3D convolutional kernel's depth is the number of channels of the image input. Setting a deeper kernel can make the network take more advantage of the temporal information in the video, while a kernel with less channels will focus more on nearby neighboring frames. With a proper design of the 3D kernel's size, the receptor field can be adjusted to concentrate on a limited speckle region for position estimation. For example, taking a $N \times H \times W$ image as input ($N$ denotes the number of consecutive frames, $H$ and $W$ denote the height and width of the image, respectively), the first 3D convolutional layers is designed to have 64 3D kernels sized $7\times7\times3$. The stride equals to 2 on the image plane with paddings and 1 along the image channels (frames) without paddings. Such design pattern can capture a relatively large receptor field on the image while uniting 3 neighboring frames each time for temporal information extraction.

During the training process, we stack a sequence of $N$ frames with height and width denoted by $H$ and $W$, respectively, to form a 3D input volume in the shape of $N \times H \times W$. Let $\{ \mathbf{\theta}_i | i=1,\ldots,N-1 \}$ denote the relative transform parameters between the neighboring frames. Instead of directly using these parameters as ground-truth labels for network training, the mean parameters
\begin{equation}
\bar{\theta} = \frac{1}{N-1} \sum_{i=1}^{N-1} \theta_i,
\end{equation}
are used for the following two reasons.
Most importantly, since the magnitude of motion between two frames is small, using the mean can effectively smooth the noise in probe motion. Another advantage in practice is that there is no need to modify the output layer every time when we change the number of input frames.
%Theoretically, taking this 3D image volume as input, the corresponding training label should be a vector $\left \{\theta_1, \theta_2,...,\theta_{N-1} \right \}$ containing the inter-frame motion between each neighboring pair. However, we argue that using the mean of these $N-1$ motion vector, denoting as $\overline{\theta_i}$, as the training label is more appropriate and efficient because: (1) Using the mean motion vector can effectively smooth the probe's unconsciously fluctuation and gives a more general motion trajectory. (2) It can be a unnecessarily complex and difficult task for the network to regress $(N-1)*6$ values when $N$ is a very large number. (3) What's worse, the output layer (fully connected layer) needs to be modified every time we switch the input format. In summary, the proposed DCL-Net takes a 3D volume shaped $N \times H \times W$ as input and estimate the average inter-frame motion $\overline{\theta_i}$ along these $N$ frames.
During the test, we slide along the video sequence with a window size $N$. The inter-frame motion of two neighboring frames is the average motion computed in all the batches.
%\blue{This is a little bit confusing to explain clearly, actually looking for suggestions...During the reconstruction, I need to generate the interframe motion for every neighboring pair. So I get the average motion of all the video segments that contain such a pair and assign to this pair as the result.} 
% vector composed of 3 translations and a unit quaternion in the order of (x, y, z, w) denoting the rotation information. A 3x3 rotation matrix can be calculated by Equation~\ref{quaternion}:
% \begin{equation} \label{quaternion}
% R=\begin{bmatrix}
% w^2+x^2-y^2-z^2 & 2xy-2wz & 2xz+2wy\\ 
% 2xy+2wz & w^2-x^2+y^2-z^2 & 2yz-2wx\\ 
% 2xz-2wy & 2yz+2wx & w^2-x^2-y^2+z^2
% \end{bmatrix}\end{equation}

\subsection{Attention Module}

% \redb{Insert a figure to show the key speckles that warrant attention.}

The attention mechanism in the deep learning models makes the CNN to focus on a specific region of an image, which carries salient information for the targeted task \cite{bahdanau2014neural}. It has led to significant improvement in various computer vision tasks such as object classification~\cite{fukui2019attention} and segmentation~\cite{oktay2018attention}. 
%These tasks have one thing in common: some specific areas of the image can dominantly contribute to the final results. 
In our 3D US volume reconstruction task, regions with strong speckle patterns for correlation are of high importance in estimating the transformations.
Thus, we introduce a self-attention block, as shown in Fig.~\ref{fig:framework}, which takes the feature maps produced by the last residual block as input and then outputs an attention map. This helps assign more weights to the highly informative regions. 
%The attention map is then multiplied with the last residual block's output and the product is passed to the next layer. The attention map's size is designed to match that of the last residual block's output, making it feasible to generate a class activation mapping for each degree-of-freedom. The visualization of a well-trained attention module's results will be displayed later in Section ~\ref{experiments}.

\subsection{Case-wise Correlation Loss}
\label{sec:correlation}

\begin{figure}[t]
	\centering
	\includegraphics[width=\textwidth]{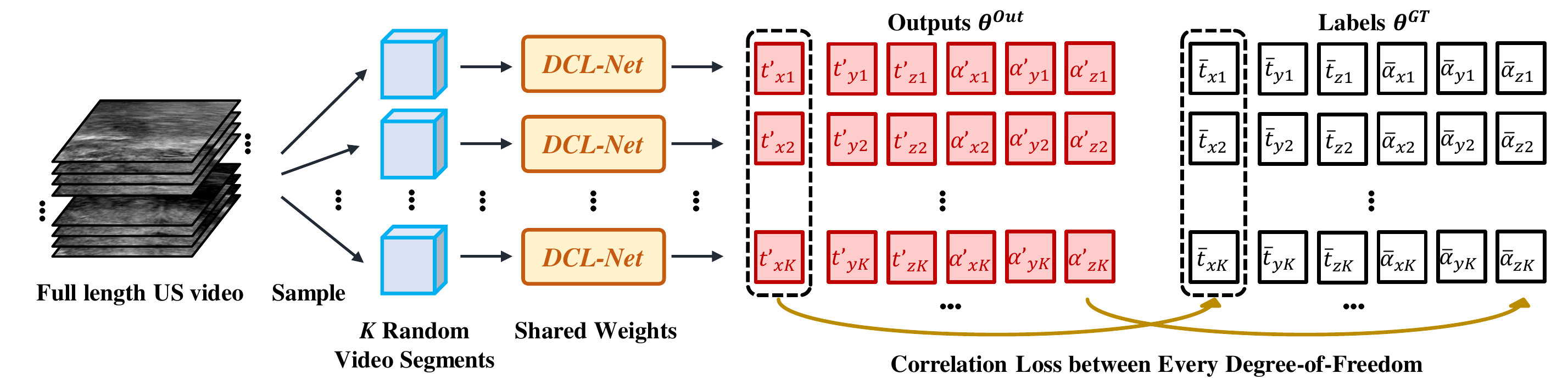}
	\caption{Overview of the case-wise correlation loss computation.} 
	\label{corrflow}
\end{figure}

%The entire DCL-Net is trained in an end-to-end fashion under the penalty of a hybrid loss function. The mean squared error (MSE) loss is the most commonly used penalty for deep regression problems. We adopt the MSE loss
%\begin{equation}
%% L_{1} = \frac{\sum_{1}^{N}Euc(P_{i}^{gt}, P_{i}^{t})}{N}
%L_{1}=\frac{1}{6K}\sum_{k=1}^{K} \sum_{d=1}^{6}(\theta_{kd}^{GT}-\theta_{kd}^{Out})^{2}
%\end{equation}
%where $K$ is the number of samples in each training batch; $\theta_{kd}^{GT}$ and $\theta_{kd}^{Out}$ are the $d$th degree of freedom in the $k$th sample of ground-truth label and network-predicted result, respectively.
%
The loss function of the proposed DCL-Net consists of two components. The first one is the mean squared error (MSE) loss, which is the most commonly used loss in deep regression problems. However, the use of MSE loss alone can lead to the smoothed estimation of the motion and thus the trained network tends to memorize the general style of how the clinicians move the probe, \emph{i.e.} the mean trajectory of the ultrasound probes. This shortcoming of the MSE loss for network training has been reported before~\cite{yang2018low,johnson2016perceptual}. To deal with problem, we introduce a case-wise correlation loss based on the Pearson correlation coefficient to emphasize the specific motion pattern of a scan.
%However, the MSE loss can easily yield a mediocre result where the output of the network tries to counter-balance all the training cases and reach a local minimum. When this happens, the network can only memorize a general style of how the clinicians move the probe while cannot flexibly react to the rapid change of moving speed and direction. \cyan{This shortcoming of the MSE loss for network training has been reported before~\cite{yang2018low,johnson2016perceptual}.}Thus, we additionally introduce a correlation loss term on the basis of Pearson Correlation Coefficient.

Fig.~\ref{corrflow} shows the workflow of calculating the case-wise correlation loss. $K$ video segments with each having $N$ frames are randomly sampled from a TRUS video.
%For every training batch, we randomly select one subject's TRUS video and randomly select $K$ segments from it, each segment contains $N$ consecutive frames. The mean motion vector of all neighboring frame-pairs within one segment is denoted as this segment's ground-truth label. In side each batch, 
The correlation coefficients between the estimated motion and the ground truth mean are computed for every degree-of-freedom and the loss is denoted as
\begin{equation}
L_{corr} = 1-\frac{1}{6}\sum_{d=1}^{6}\frac{Cov\left (\bar{\theta}_{d}^{GT}, \bar{\theta}_{d}^{Out} \right )}{\sigma \left (\bar{\theta}_{d}^{GT}  \right )\sigma \left ( \bar{\theta}_{d}^{Out} \right )},
\end{equation}
where $Cov$ gives the covariance and $\sigma$ calculates the standard deviation.
%where $\theta_{d}^{GT}$ and $\theta_{d}^{Out}$ denotes the vector of $d$-th degrees-of-freedom of the ground-truth label and the network's output, respective. In this case, $\theta_{d}^{GT}$ and $\theta_{d}^{Out}$ both contain 20 values since there are 20 segments selected from one subject. Training under the case-wise correlation loss can enforce the network to learning discriminative features that decides the spatial relationship between image pairs. In a nutshell, the prediction of the network should follow the same trend as the ground-truth motion: larger inter-frame motion should be matched with larger output estimation, and vice versa. 
The total loss is the summation of the MSE loss and the case-wise correlation loss.
% with a weighted ratio $\lambda$:
% \begin{equation}
% L_{2}=L_{1}+\lambda L_{2}
% \end{equation}

\section{Experiments and Results}
\label{experiments}
% \blue{(Organizing. I got the results now and am trying to deliver them in good-looking formats...)}

%\noindent \textbf{Settings:} 
For the experiments performed in this study, a total of 640 TRUS scanning videos (640 patients) from the Nation Institute of Health (NIH) were acquired from IRB-approved clinical trial. During an intervention, a physician used an end-firing transrectal ultrasound probe to acquire axial images by steadily sweeping through the prostate from base to apex. The positioning information given by an electromagnetic tracking device serves as the ground truth label in our training phase. The dataset is split into 500, 70 and 70 cases as training, validation and testing, respectively. Our network is trained for 300 epochs with batch size $K=20$ using Adam optimizer~\cite{kingma2014adam} with initial learning rate of 5$\times 10^{-5}$, which decays by 0.9 after 5 epochs. 
Since the prostate US image only takes a relative small part of each frame, each frame is cropped without exceeding the imaging field and then resized to $224 \times 224$ to fit the design of ResNexts~\cite{xie2017aggregated}.  We implemented the DCL-Net using the publicly available Pytorch library~\cite{pytorch}. The entire training phase of the DCL-Net takes about 4h, taking 5 frames as input. During testing, it takes about 2.58s to produce all the transformation matrix of an US video with 100 frames.

%\noindent \textbf{Evaluation metric:} 
Two evaluation metrics are used for performance evaluation. The first one is the average distance between all the corresponding frame corner-points throughout a video. This distance error reveals the difference in speed and orientation variations across the entire video. The other one is the final drift~\cite{prevost20183d}, which is the distance between the center points of the transformed end frames of a video segment using the EM tracking data and our DCL-Net estimated motion.
%(1) We get the orientation vectors of the last reconstructed frame of both ground-truth and our prediction. The distance between these two vectors are denoted as final drift~\cite{prevost20183d} which measures the ending deviation. (2) We compute the mean distance of every corresponding corner-points pairs of every frame of the ground-truth and our prediction. This distance error reveals the difference in speed and orientation variations across the video, from the initial frame to the end. 

\subsection{Parameter Setting}

\begin{figure}[t]
	\subfloat{\includegraphics[width=.45\textwidth]{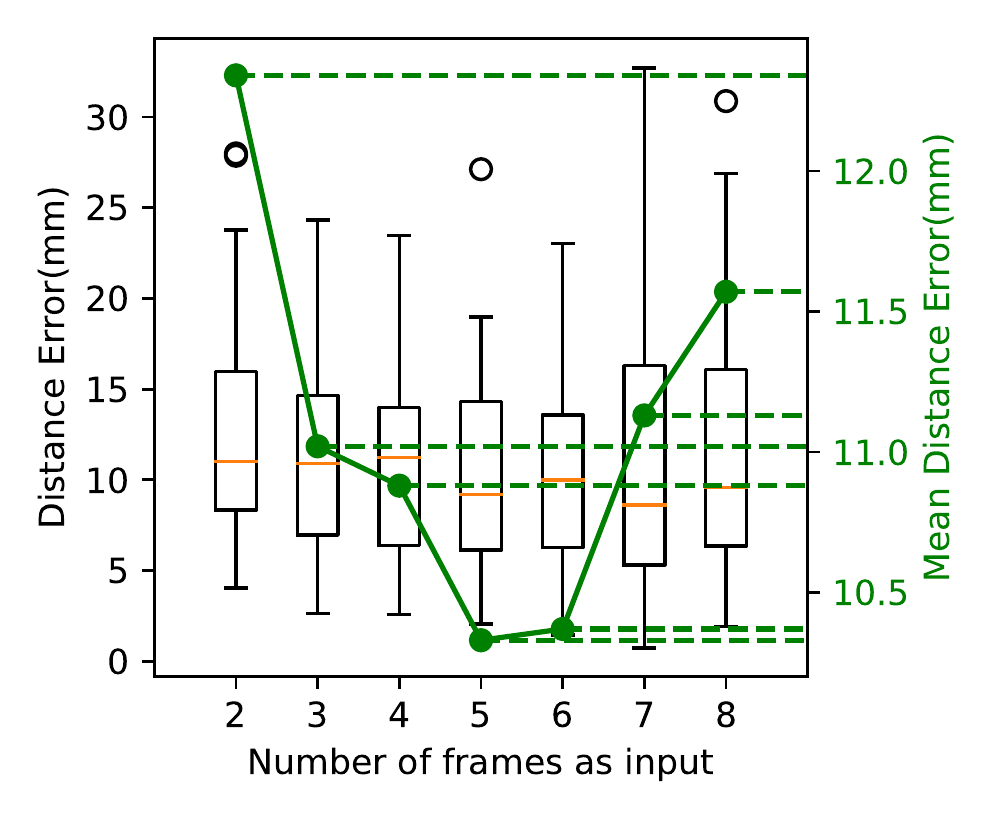}}
	\subfloat{\includegraphics[width=.55\textwidth]{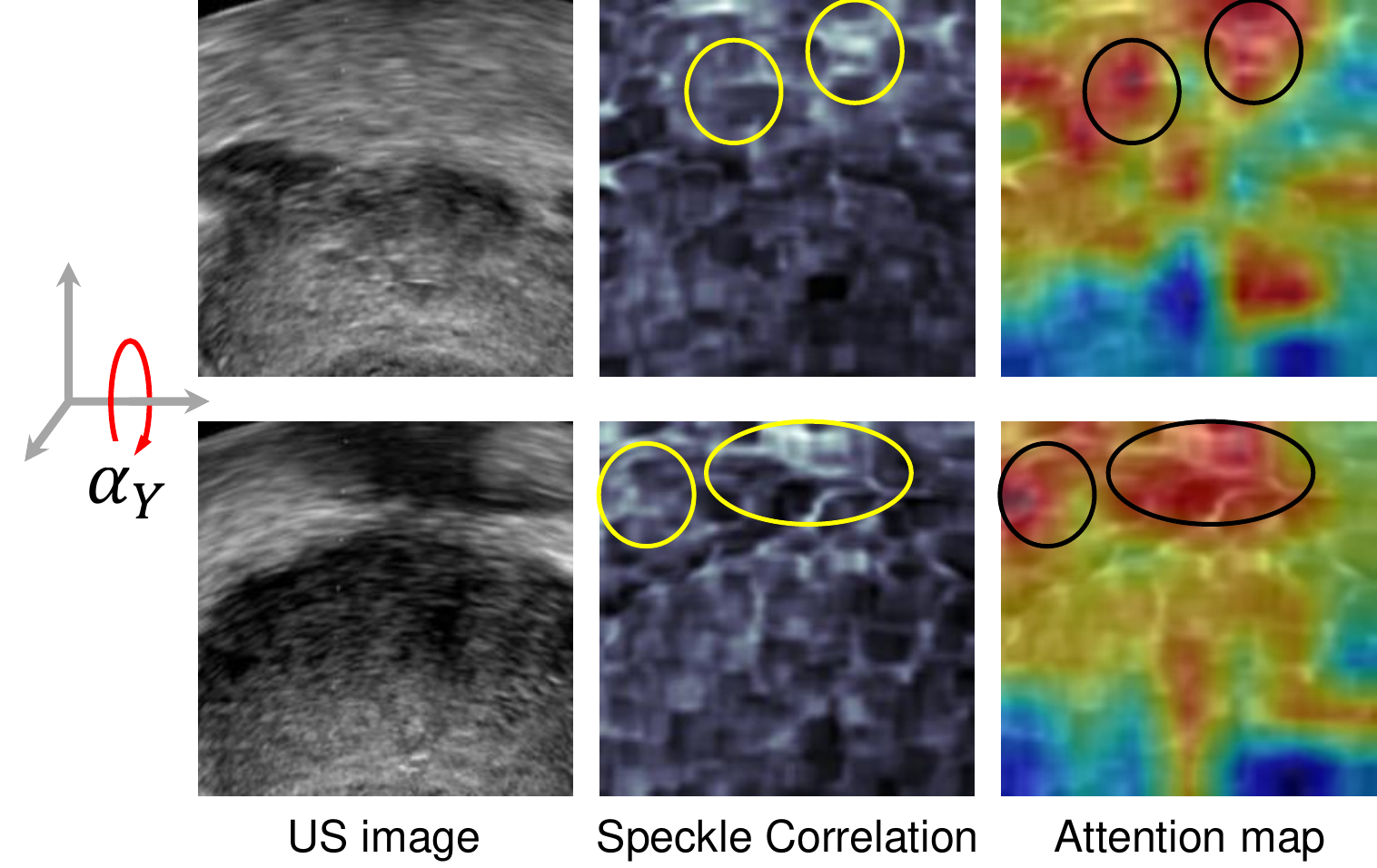}}
	\caption{(Left) Effect of number of input frames. Green curve shows the mean distance error of each box. (Right) Visualization of two attention maps regarding rotation around the Y-axis.} \label{fig:nsat}
\end{figure}

We first performed experiments to determine an optimal $N$ number of frames for each video segment. The left panel of Fig.~\ref{fig:nsat} shows how the overall reconstruction performance varies as the number of consecutive frames changes. There is a decrease then increase in the error, with neighboring frame number equaling to 5 or 6 has similarly the best performance. According to our paired t-test, the calculated p-value is smaller than the confidence level of 0.05, indicating the result using 5 frames is significantly better than that using only 2 frames. The explanation is that the network takes advantage of the rich contextual information along the time-series and produces more stable trajectory estimation.

The right panel of Fig.~\ref{fig:nsat} visualizes two example attention maps. The first image column shows the cropped US images. The second column is the speckle correlation map~\cite{chang20033} between a US image and its following neighboring frame. Inside this speckle correlation map, brighter the area, longer the elevational distance to the next frame. Such pattern with dark areas at the bottom and brighter on the upper part is consistent with our TRUS scanning protocol, as there is less motion around the tip of the ultrasound probe. The third column shows the attention map regarding the rotation $\alpha_y$ around the Y-axis, which also indicates part of the out-of-plane rotation. The attention maps have strong activation at the bright speckle correlation regions, indicating that the attention module helps the network to focus on speckle-rich areas for better reconstruction.

\subsection{Reconstruction Performance and Discussions}
% \usepackage{multirow}
\begin{table}[t]
\caption{Performance of different methods on the EM-tracking dataset.}
\centering
\begin{tabular}{l|cccc|cccc}\hline
\multicolumn{1}{c|}{\multirow{2}{*}{Methods}}                                 & \multicolumn{4}{c|}{Distance Error (mm)}                            & \multicolumn{4}{c}{Final Drift (mm)}                                \\ \cline{2-9} 
\multicolumn{1}{c|}{}                                                         & Min           & Median        & Max            & Average       & Min           & Median        & Max           & Average        \\\hline 
Linear Motion                                                                 & 7.17          & 19.73         & 60.79          & 22.53         & 12.53         & 37.15         & 114.02        & 42.62          \\
Decorrelation~\cite{chang20033}    & 9.62          & 17.58         & 56.72          & 18.89         & 15.32         & 38.45         & 104.13        & 38.26          \\
2D CNN~\cite{prevost20183d}                                                           & 5.66          & 15.8          & 43.35          & 17.42         & 7.05          & 23.13         & 68.87         & 26.81          \\
3D CNN (NS2)~\cite{xie2017aggregated}                                                                 & 2.38          & 10.14         & 31.34          & 12.34         & 1.42          & 19.08         & 68.61         & 21.74          \\
% 3D CNN(NS5)                                                                 & 2.13          & 11.22         & \textbf{23.09}          & 11.6          & 2.36          & 17.64         & 57.63         & 19.65          \\
% 3D CNN(NS5)+Attn                                                          & 2.06          & 10.78         & 23.65          & 11.17         & 2.82          & \textbf{16.96}         & 56.84         & 19.06          \\
Our DCL-Net  & \textbf{1.97} & \textbf{9.15} & \textbf{27.03} & \textbf{10.33} & \textbf{1.09} & \textbf{17.40} & \textbf{55.50} & \textbf{17.39}\\\hline
\end{tabular}
\label{tabcompare}
\end{table}

Table~\ref{tabcompare} summarizes the overall comparison of the proposed DCL-Net against other existing methods. The approach of ``Linear Motion'' means that we first calculate the mean motion vector of the training set and then apply this fixed vector to all the testing cases. The approach of ``Decorrelation'' is based on the speckle decorrelation algorithm presented in \cite{chang20033}. ``2D CNN'' refers to the method presented by Prevost \etal~\cite{prevost20183d}. ``3D CNN'' is the vanilla ResNext~\cite{xie2017aggregated} architecture taking only two slices as input.

It can be seen from Table~\ref{tabcompare} that our proposed DCL-Net outperforms all the other methods. Paired t-test was performed and the performance improvement made by DCL-Net is significant in all the cases with $p$-value$<$0.05. It is worth noting that although the average distance error of 10.33mm achieved DCL-Net is still a large error, this is the best performance on real clinical data instead of phamtom studies. The performance of the state-of-the-art 2D-CNN method reproduced in our experiments has consistent performance compared to the accuracy reported in the paper \cite{prevost20183d}. It is a challenging task to reconstruct 3D US volume using these freehand TRUS scans and we have been making significant progress in this important area.
%We reproduce the experiment of 2D-CNN proposed in  and the yieldings on our dataset is consistent with what they report originally. 
%The statistical tests have been carried out between each baseline method and our DCL-Net. All the calculated p-value is smaller than the significance level of 0.05, indicating that our approach is significantly better than all the listed baseline methods.  
% \redb{Discuss the numbers in your table. 10.33 sounds like a very large number. What does it tell others? What are the numbers in Prevost's paper? You can emphasize that this is real clinical data and thus more challenging. }

%\subsection{Discussions}

\begin{figure}[t]
	\centering
	\includegraphics[width=\textwidth]{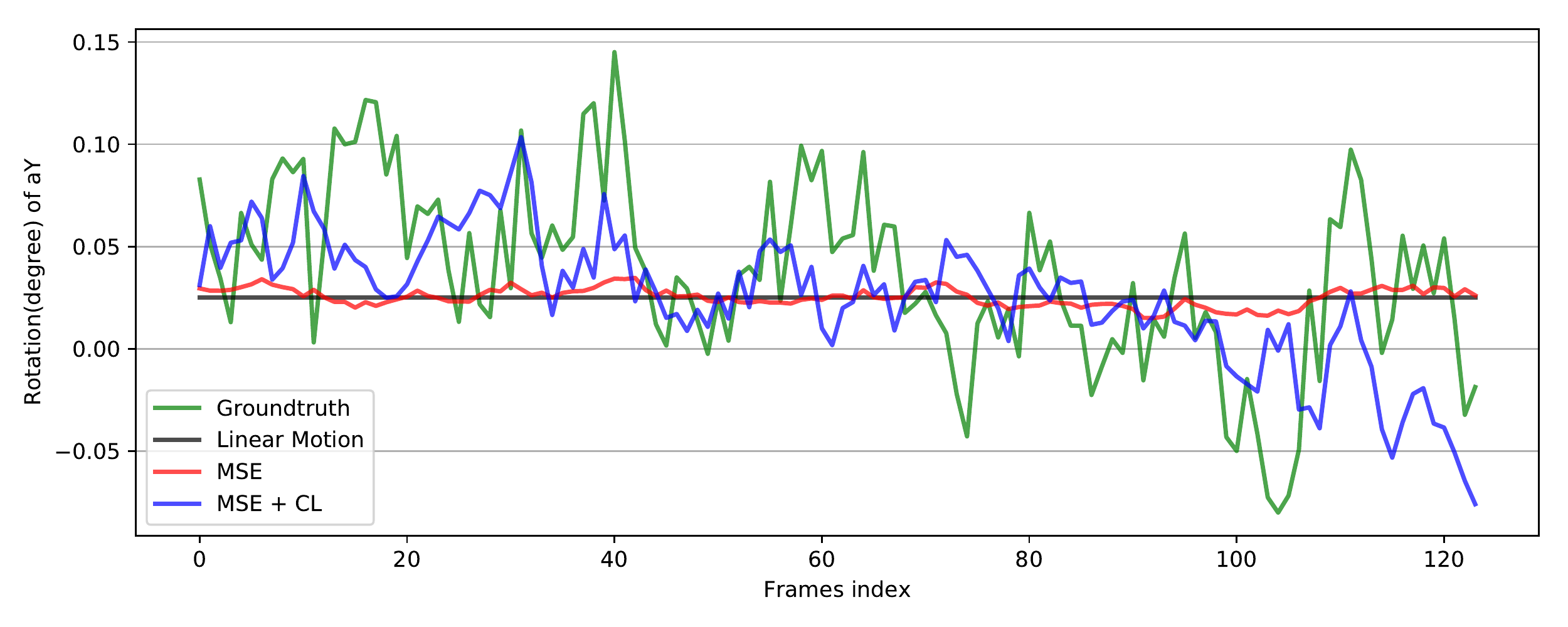}
	\caption{Predicted rotation $\alpha_y$ on one video sequence with different methods. Applying correlation loss makes our prediction (blue line) more sensitive to the strongly varying speed of the ground-truth (green line).} 
	\label{fig:corrstudy}
\end{figure}

\begin{figure}[t]
	\centering
	\includegraphics[width=\textwidth]{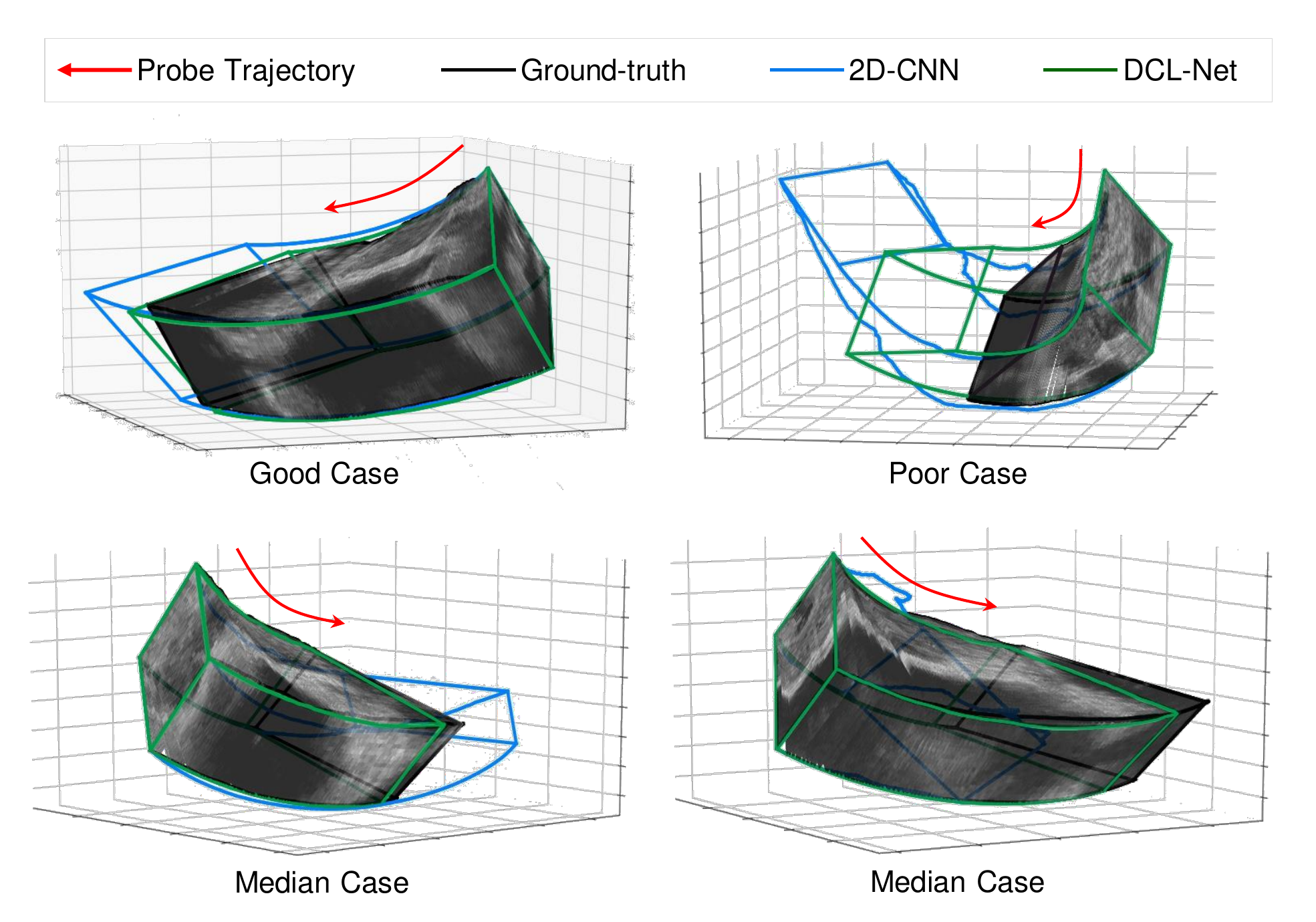}
	\caption{Comparison of the US volume reconstruction results of four cases with different qualities.}
	\label{fig:recon}
\end{figure}

Next, we demonstrate the effectiveness of incorporating case-wise correlation loss into the network training. Fig.~\ref{fig:corrstudy} shows the prediction of $\alpha_y$ along a video sequence. We can observe that the network trained with MSE loss can only produce mediocre results (red line) which are nearly constant, showing almost no sensitivity to the change in speed and orientation. Its prediction wonders around the linear motion (black line) which represents the mean value of the probe's trajectory. The correlation coefficients of all testing cases show a mean of $0.09 \pm 0.03$ which represents little correlation. This indicates that using MSE alone makes the network memorizing the general style of the US probe motion trajectory and fails to produce valid prediction based on image contents. By incorporating the correlation loss (CL) into the loss function (blue line), the correlation coefficients of all testing cases have a mean of $0.21 \pm 0.09$, representing weak correlation. Based on a paired t-test with $\alpha=0.05$, this is found to be significantly better than the previous results. Intuitively, the network's prediction reacts more sensitively to the variation of the probe's real translation and rotation (green line). 

Last but not the least, we report the volume reconstruction results using four testing cases with different reconstruction qualities as shown in Fig.~\ref{fig:recon}. One good case, one bad case, and two median cases are included to offer a complete view of the performance. To reduce the clutter in the figure, we only show the comparison between our DCL-Net, the 2D-CNN~\cite{prevost20183d} and the ground-truth. While producing competitive performance, the 2D-CNN method is less sensitive to the speed variations of US probe and the estimated trajectory has noisy vibration. The results sometimes even severely deviate from the ground-truth. Our proposed DCL-Net shows a much smoother trajectory estimation thanks to the contextual information provided by video segments. %Also, our method can better match the speed of US probe in majority of cases. 

\section{Conclusions}
\label{Conclusion}
This paper introduced a sensorless freehand 3D US volume reconstruction method based on deep learning. The proposed DCL-Net can well extract the information among multiple US frames to improve the US probe trajectory estimation. Experiments on a well-sized EM-tracked ultrasound dataset demonstrated that the proposed DCL-Net has benefited from the contextual learning and showed superior performance when compared to other existing methods. Further experiments on the ultrasound videos with different scanning protocols will be studied in our future work.

\section{Acknowledgements}
\label{Acknowledgements}
This work was partially supported by National Institute of Biomedical Imaging and Bioengineering (NIBIB) of the National Institutes of Health (NIH) under awards R21EB028001 and R01EB027898, and through an NIH Bench-to-Bedside award made possible by the National Cancer Institute.

\bibliographystyle{splncs04} 
\bibliography{references}

\end{document}